\documentclass[letterpaper,journal,table]{IEEEtran}

\usepackage{amsmath,amsfonts,amssymb}

\usepackage{graphicx}
\usepackage{svg}
\usepackage[caption=false,font=normalsize,labelfont=sf,textfont=sf]{subfig}
\usepackage{subcaption}
\usepackage{float}
\usepackage{stfloats}
\usepackage{wrapfig}

\usepackage{array}
\usepackage{booktabs}
\usepackage{multirow}
\usepackage{tabularx}
\usepackage{threeparttable}
\usepackage{makecell}
\usepackage{xcolor}

\usepackage{algorithm}
\usepackage{algorithmic}

\usepackage{hyperref}
\usepackage{url}

\usepackage[
    style=ieee,
    natbib=true,
    citestyle=numeric,
    sorting=none,
    doi=false,
    isbn=false,
    url=false]{biblatex}

\usepackage[inter-unit-product =\cdot]{siunitx}
\usepackage{cleveref}

\usepackage{textcomp}
\usepackage{verbatim}
\usepackage{xspace}
\usepackage{bbm}

\makeatletter
\let\NAT@parse\undefined
\makeatother

\definecolor{myred}{RGB}{191, 3, 3}
\definecolor{mydarkblue}{RGB}{0, 20, 115}
\definecolor{mydarkgreen}{RGB}{0, 139, 69}
\definecolor{mygreen2}{RGB}{0, 205, 0}
\definecolor{mybrown}{RGB}{139, 69, 19}

\hypersetup{
    colorlinks=true,
    linkcolor=magenta,
    urlcolor=magenta,
    citecolor=mygreen2,
}

\Crefname{asm}{Assumption}{Assumption}

\newcommand{\ULC}{\textcolor{myred}{ULC}\xspace}
\newcommand{\ci}[1]{\scriptsize{$\pm$#1}}

\begin{document}

\title{
    \ULC: A \textcolor{myred}{U}nified and Fine-Grained \textcolor{myred}{C}ontroller for Humanoid \textcolor{myred}{L}oco-Manipulation
}

\author{Wandong Sun$^{1*}$, Luying Feng$^{2*}$, Yang Liu$^{1*}$, Baoshi Cao$^{1}$, Yaochu Jin$^{2}$\textsuperscript{\textdagger}, Zongwu Xie$^{1}$\textsuperscript{\textdagger}%
    \thanks{$^{*}$These authors contributed equally to this work}%
    \thanks{\textdagger\ denotes the corresponding author}%
    \thanks{$^{1}$State Key Laboratory of Robotics and Systems, Harbin Institute of Technology.
    }%
    \thanks{$^{2}$School of Engineering, Westlake University.
    }%
    \vspace{-8pt}
}

\pagestyle{empty}

\twocolumn[{
            \renewcommand\twocolumn[1][]{#1}
            \maketitle
            \thispagestyle{empty}
            \begin{center}
                \vspace{-20pt}
                \captionsetup{type=figure}
                \includegraphics[width=\textwidth]{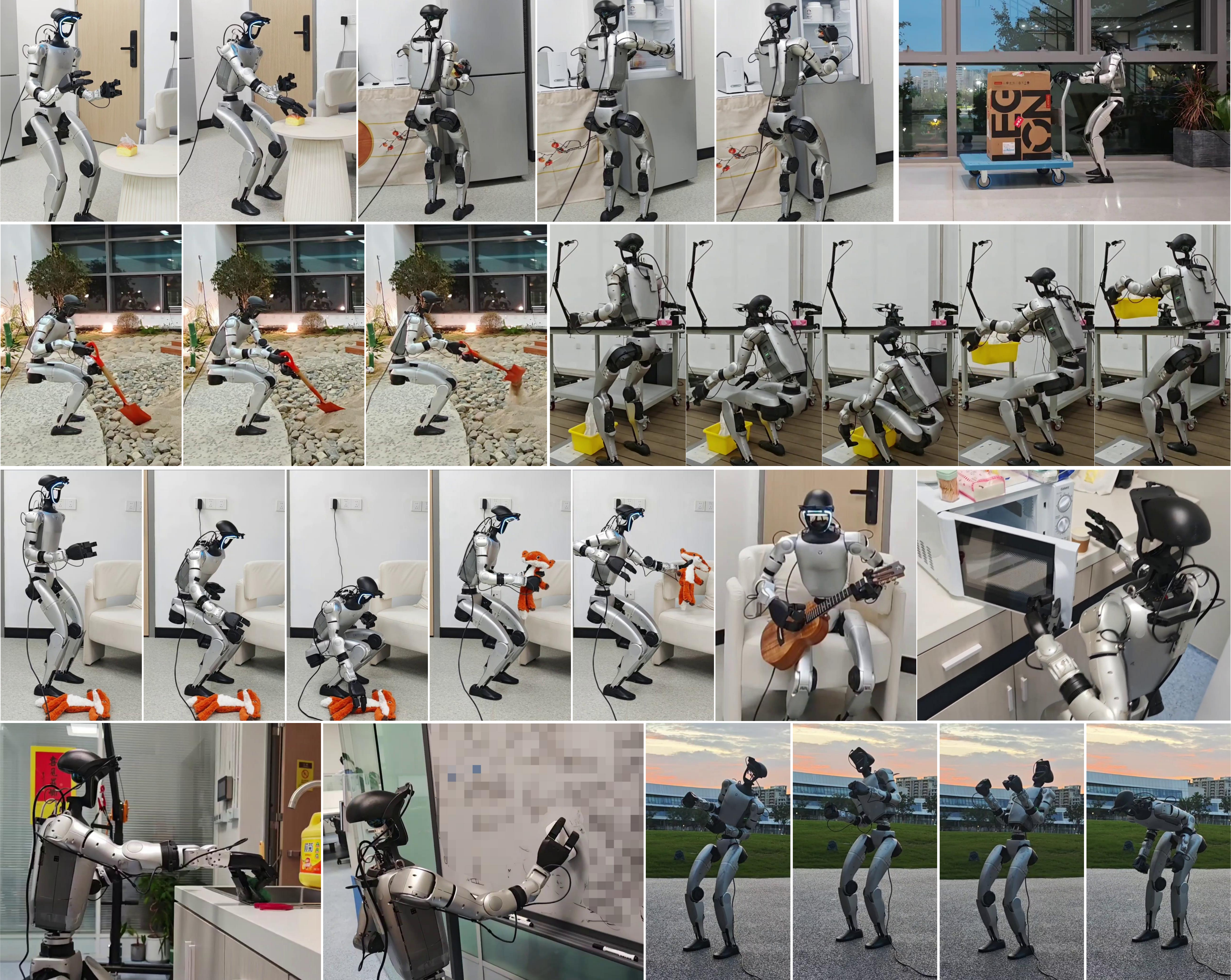}
                \captionof{figure}{Diverse loco-manipulation capabilities enabled by \ULC. The humanoid robot demonstrates various coordinated whole-body actions, including picking up bread from a table and placing it in a refrigerator, pushing a cart with coordinated locomotion, squatting to shovel sand from the ground, lifting boxes from the floor to table height with dual-arm coordination, picking up dolls from the ground with hand switching and placing them on a sofa, sitting and playing ukulele with fine motor control, placing items in a microwave with precise manipulation, cleaning kitchen surfaces with wiping motions, erasing blackboards with arm coordination, and performing torso rotation in outdoor environments.}
                \label{fig:demo}
            \end{center}
        }]

\begin{abstract}
    Loco-manipulation for humanoid robots aims to enable robots to integrate
mobility with upper-body tracking capabilities. Most existing approaches adopt
hierarchical architectures that decompose control into isolated upper-body
(manipulation) and lower-body (locomotion) policies. While it reduces training
complexity, this decomposition inherently limits coordination between
subsystems and contradicts the unified whole-body control exhibited by humans.
We demonstrate that a single unified policy can achieve a combination of
tracking accuracy, large workspace, and robustness for humanoid
loco-manipulation. We propose a \textcolor{myred}{U}nified
\textcolor{myred}{L}oco-Manipulation \textcolor{myred}{C}ontroller (\ULC), a
single-policy framework that simultaneously tracks root velocity, root height,
torso rotation, and dual-arm joint positions in an end-to-end manner,
demonstrating the feasibility of unified control without sacrificing the
performance. We achieve this unified control through integrating a set of key
technologies, including sequential skill acquisition for progressive learning
complexity, residual action modeling for fine-grained control adjustments,
command polynomial interpolation for smooth motion transitions, random delay
release for robustness to deploy variations, load randomization for
generalization to external disturbances, and center of mass tracking for
providing explicit policy gradients to maintain stability. We validate our
method on the Unitree G1 humanoid robot with a three-degrees-of-freedom waist.
Compared with the state-of-the-art, \ULC shows better tracking performance than
disentangled methods and demonstrates larger workspace coverage. The unified
dual-arm tracking enables precise manipulation under external loads while
maintaining coordinated whole-body control for complex loco-manipulation tasks.
The code and videos are available on our project
website at \url{https://hellod035.github.io/ULC/}.

\end{abstract}

\section{INTRODUCTION}
\label{sec:intro}
\begin{table*}[htbp]
      \centering
      \small
      \vspace{5pt}
      \renewcommand{\arraystretch}{1.2}
      \begin{tabular}{l c c c c c c c c}
            \toprule
            \textbf{Method}                                       & \textbf{Architecture} & \textbf{Legs}           & \textbf{Torso Yaw}      & \textbf{Torso Pitch}    & \textbf{Torso Roll}     & \textbf{Dual Arms}      & \textbf{Workspace} & \textbf{Precision} \\
            \midrule
            HOMIE~\cite{benHOMIEHumanoidLocoManipulation2025}     & Decoupled             & \cellcolor{blue!20}RL-1 & \cellcolor{orange!20}PD & \cellcolor{gray!20}-    & \cellcolor{gray!20}-    & \cellcolor{orange!20}PD & Medium             & Medium             \\
            FALCON~\cite{zhangFALCONLearningForceAdaptive2025}    & Decoupled             & \cellcolor{blue!20}RL-1 & \cellcolor{blue!20}RL-1 & \cellcolor{blue!20}RL-1 & \cellcolor{blue!20}RL-1 & \cellcolor{red!20}RL-2  & Medium             & High               \\
            JAEGER~\cite{dingJAEGERDualLevelHumanoid2025}         & Decoupled             & \cellcolor{blue!20}RL-1 & \cellcolor{blue!20}RL-1 & \cellcolor{gray!20}-    & \cellcolor{gray!20}-    & \cellcolor{red!20}RL-2  & Medium             & High               \\
            AMO~\cite{liAMOAdaptiveMotion2025}                    & Decoupled             & \cellcolor{blue!20}RL   & \cellcolor{blue!20}RL   & \cellcolor{blue!20}RL   & \cellcolor{blue!20}RL   & \cellcolor{orange!20}PD & Large              & Medium             \\
            SoFTA~\cite{liHoldMyBeer2025}                         & Decoupled             & \cellcolor{blue!20}RL-1 & \cellcolor{blue!20}RL-1 & \cellcolor{gray!20}-    & \cellcolor{gray!20}-    & \cellcolor{red!20}RL-2  & Medium             & Medium             \\
            R$^2$S$^2$~\cite{zhangUnleashingHumanoidReaching2025} & Unified               & \cellcolor{purple!20}RL & \cellcolor{purple!20}RL & \cellcolor{purple!20}RL & \cellcolor{purple!20}RL & \cellcolor{purple!20}RL & Medium             & Medium             \\
            \midrule
            \textcolor{myred}{\textbf{ULC}} (Ours)                & Unified               & \cellcolor{purple!20}RL & \cellcolor{purple!20}RL & \cellcolor{purple!20}RL & \cellcolor{purple!20}RL & \cellcolor{purple!20}RL & Large              & High               \\
            \bottomrule
      \end{tabular}
      \caption{Comparison of humanoid loco-manipulation controllers. Colors indicate control types: \textcolor{blue!50}{Blue}/\textcolor{red!50}{Red}: RL, \textcolor{orange!50}{Orange}: PD, \textcolor{purple!50}{Purple}: Unified RL, \textcolor{gray!50}{Gray}: Not controlled.}
      \label{tab:loco_manipulation_comparison}
      \vspace{-10pt}
\end{table*}

Humanoid robots, with their human-like morphology, represent a promising
paradigm for versatile systems that can operate in human-centric environments.
Recent advances have significantly improved
locomotion~\cite{sunLearningPerceptiveHumanoid, heASAPAligningSimulation2025,
      allshireVisualImitationEnables2025, ben2025gallant, packer2019assessinggeneralizationdeepreinforcement, zhuangRobotParkourLearning2023, zhuangHumanoidParkourLearning2024a, zhengFLARERobotLearning2025} and
manipulation~\cite{qiuHumanoidPolicyHuman2025,
      linSimtoRealReinforcementLearning2025,
      chenDexForceExtractingForceinformed2025, intelligence2025pi05visionlanguageactionmodelopenworld, nvidia2025gr00tn1openfoundation}. These results are often achieved by
coupling high-level decision-making—via Imitation Learning
(IL)~\cite{chi2024diffusionpolicyvisuomotorpolicy,
      zhao2023learningfinegrainedbimanualmanipulation} or Vision-Language-Action
(VLA) models~\cite{black2024pi0visionlanguageactionflowmodel,
      intelligence2025pi05visionlanguageactionmodelopenworld,
      nvidia2025gr00tn1openfoundation}—with Loco-Manipulation Controllers
(LMCs)~\cite{benHOMIEHumanoidLocoManipulation2025, liAMOAdaptiveMotion2025}
that map commands to whole-body motions.

An effective LMC should translate motion commands into joint-level actions with
minimal tracking error while maintaining dynamic stability. However, LMC design
involves key trade-offs. The command space (e.g., joint positions, Cartesian
poses, root velocity, root height) affects feasibility, conflicts across
objectives, and how fully the robot can exploit its kinematic and dynamic
range~\cite{heHOVERVersatileNeural2024}. The control architecture also matters:
whole-body controllers~\cite{heOmniH2OUniversalDexterous2024,
      heHOVERVersatileNeural2024, fuHumanPlusHumanoidShadowing2024, zhuangEmbraceCollisionsHumanoid2025, zhangHuBLearningExtreme2025,
      heASAPAligningSimulation2025, videomimic} offer strong
coordination but are harder to train, whereas decoupled upper/lower-body
designs~\cite{benHOMIEHumanoidLocoManipulation2025,
      dingJAEGERDualLevelHumanoid2025, zhangFALCONLearningForceAdaptive2025} simplify
learning but can weaken coordination. Finally, training data sources impose
different biases: motion capture provides realistic motion but can be noisy,
kinematically infeasible, and biased~\cite{heOmniH2OUniversalDexterous2024,
      heHOVERVersatileNeural2024, fuHumanPlusHumanoidShadowing2024, gmt, any2track, resmimic,
      beyondmimic}; procedurally
sampled commands improve coverage but are often limited (especially for legs)
due to humanoid stability constraints~\cite{jeneltenDTCDeepTracking2024a, xueUnifiedGeneralHumanoid2025,
      zhangFALCONLearningForceAdaptive2025, benHOMIEHumanoidLocoManipulation2025}.

These choices ultimately determine deployability: command design must balance
expressiveness and feasibility, architectures must balance coordination and
training cost, and data generation must balance realism and coverage.

To address these issues, we propose a \textcolor{myred}{U}nified
\textcolor{myred}{L}oco-Manipulation \textcolor{myred}{C}ontroller (\ULC)
trained with massively parallel reinforcement learning to track procedurally
sampled commands including root velocity, root height, torso orientation, and
arm joint positions. By simplifying leg commands compared to
motion-capture-driven approaches, we improve coverage of the feasible command
space while preserving whole-body coordination. We enable single-model
multi-task tracking via (i) feasibility-aware command space design, (ii)
progressive curriculum learning, (iii) residual action
modeling~\cite{silver2018residual, johannink2019residual,
      heASAPAligningSimulation2025} to improve tracking precision, and (iv)
sequential skill acquisition~\cite{luoPerpetualHumanoidControl2023} to reduce
catastrophic forgetting. For deployment-realistic command generation, we
combine fixed-interval random sampling with fifth-degree polynomial
interpolation~\cite{guAdvancingHumanoidLocomotion2024,
      guHumanoidGymReinforcementLearning2024} and introduce stochastic command
release that probabilistically buffers/releases commands to emulate deployment
variations while keeping commands feasible. To improve robustness under varying
payloads, we add center of mass tracking
rewards~\cite{zhangHuBLearningExtreme2025} encouraging the center of mass
projection to remain within the support polygon. Experiments in simulation and
on hardware show state-of-the-art tracking performance, workspace coverage, and
robustness; ablations verify the role of each component. Our contributions are:

\begin{itemize}
      \item A unified framework with feasibility-aware command space design and progressive
            curriculum learning for multi-task loco-manipulation.
      \item Deployment-realistic training with stochastic command release and explicit
            balance optimization for sim-to-real transfer and payload robustness.
      \item Extensive experiments demonstrating improved tracking performance and
            generalization across diverse tasks.
\end{itemize}


\section{RELATED WORK}
\label{sec:related_work}

\subsection{Humanoid Loco-Manipulation Controller}
Humanoid loco-manipulation control confronts fundamental challenges in
coordinating locomotion with manipulation while maintaining tracking accuracy
and system robustness \cite{zhangFALCONLearningForceAdaptive2025,
      benHOMIEHumanoidLocoManipulation2025, liAMOAdaptiveMotion2025}.

Traditional approaches frequently employ decoupled control strategies that
isolate leg and arm movements to reduce training complexity. Representative
implementations integrate reinforcement learning for leg control with PD
controllers for arms \cite{benHOMIEHumanoidLocoManipulation2025} while
demonstrating suboptimal arm tracking under gravitational loads. Force
adaptation challenges are tackled through joint upper-body policy training with
force curriculum \cite{zhangFALCONLearningForceAdaptive2025}. Dual-level
control architectures implement separate upper/lower body controllers capable
of root velocity tracking and fine-grained joint control
\cite{dingJAEGERDualLevelHumanoid2025}, yet remain dependent on motion
retargeting that may introduce artifacts. Hand stabilization during locomotion
is achieved through multi-frequency frameworks using distinct upper/lower-body
agents \cite{liHoldMyBeer2025}. A new perspective switching dual-arm input
source via binary flags \cite{xueUnifiedGeneralHumanoid2025} demonstrates
conceptual advancement, but exhibits under-refined dual-arm tracking under PD
control. Skill-space methodologies enable extended mobility through primitive
ensembling \cite{zhangUnleashingHumanoidReaching2025}. However, none of the
above methods completely frees up the working space of trunk rotation.
Hierarchical designs combining trajectory optimization with reinforcement
learning \cite{liAMOAdaptiveMotion2025} show performance improvements at the
cost of computational overhead, while accurate dual-arm tracking remains
unintegrated.


\begin{table}[t]
      \centering
      \small
      \vspace{5pt}
      \begin{tabular}{l c c}
            \toprule
            \textbf{Parameter}   & \textbf{Unit} & \textbf{Range}      \\
            \midrule
            Linear Velocity X    & m/s           & [-0.45, 0.55]       \\
            Linear Velocity Y    & m/s           & [-0.45, 0.45]       \\
            Angular Velocity Z   & rad/s         & [-1.2, 1.2]         \\
            Root Height          & m             & [0.3, 0.75]         \\
            Torso Rotation Yaw   & rad           & [-2.62, 2.62]       \\
            Torso Rotation Roll  & rad           & [-0.52, 0.52]       \\
            Torso Rotation Pitch & rad           & [-0.52, 1.57]       \\
            Arm Joint Positions  & -             & Robot Design Limits \\
            \bottomrule
      \end{tabular}
      \caption{Command space specifications.}
      \label{tab:command_space}
      \vspace{-15pt}
\end{table}

\subsection{Humanoid Whole-Body Tracking}
Recent significant strides in humanoid whole-body motion tracking now enable
robots to reproduce complex human motions using diverse datasets
\cite{zhuangEmbraceCollisionsHumanoid2025, zhangHuBLearningExtreme2025,
      heASAPAligningSimulation2025, videomimic, gmt, any2track, resmimic,
      beyondmimic}, though fundamental challenges persist regarding morphological
differences, noise handling, and sim-to-real transfer.

Recent innovations demonstrate diverse pathways for advancement. When training
on large-scale motion capture datasets \cite{heOmniH2OUniversalDexterous2024},
whole-body dexterous manipulation policies emerge yet suffer from diluted
learning of extreme motions. For data quality enhancement, teacher-student
distillation and dataset selection techniques
\cite{jiExBody2AdvancedExpressive2024} mitigate issues but introduce
performance gaps between teacher and student policies. Extreme motion
reproduction is achievable through advanced processing pipelines
\cite{zhangHuBLearningExtreme2025}, albeit with limited transferability across
motion sequences. Meanwhile, visual imitation from video demonstrations
\cite{videomimic} reduces motion capture dependency with video-to-motion
techniques. General motion tracking frameworks \cite{gmt} and robust tracking
methods \cite{any2track} advance whole-body control capabilities. Residual
learning approaches \cite{resmimic} and diffusion-based methods
\cite{beyondmimic} bridge the gap from motion tracking to versatile control.



\section{\textcolor{myred}{U}NIFIED \textcolor{myred}{L}OCO-MANIPULATION \textcolor{myred}{C}ONTROL}
\label{sec:ulc}
We present \ULC, a \emph{unified} and \emph{fine-grained} controller for
humanoid loco-manipulation that leverages massive parallel reinforcement
learning to train a single policy from scratch. Our framework systematically
addresses the fundamental challenges of high-dimensional exploration and skill
coordination through four key technical innovations:

\begin{enumerate}
    \item \textbf{Sequential skill acquisition} with adaptive curriculum;
    \item \textbf{Command interpolation} with stochastic delay modeling;
    \item \textbf{Load generalization} through dynamic mass distribution and
          center of mass tracking;
    \item \textbf{Residual action modeling} for stable training and precise upper body tracking.
\end{enumerate}

\subsection{Problem Formulation}

We formulate the humanoid loco-manipulation task as a goal-conditioned Markov
Decision Process (MDP) $\mathcal{M} = \langle \mathcal{S}, \mathcal{A},
    \mathcal{G}, P, R, \gamma \rangle$. The policy $\pi_{\boldsymbol{\theta}}:
    \mathcal{S} \times \mathcal{G} \rightarrow \Delta(\mathcal{A})$ maps
state-command observations to a Gaussian distribution over actions:
\begin{equation}
    \pi_{\boldsymbol{\theta}}(\boldsymbol{a}_t | \boldsymbol{s}_t, \boldsymbol{g}_t) = \mathcal{N}(\boldsymbol{\mu}_{\boldsymbol{\theta}}(\boldsymbol{s}_t, \boldsymbol{g}_t), \boldsymbol{\Sigma}_{\boldsymbol{\theta}}(\boldsymbol{s}_t, \boldsymbol{g}_t))
\end{equation}
The observation
\begin{equation}
    \boldsymbol{o}_{prop}^{(t)} = [\boldsymbol{q}_{joint}^{(t)}, \dot{\boldsymbol{q}}_{joint}^{(t)}, \boldsymbol{\omega}_{base}^{(t)}, \boldsymbol{g}_{proj}^{(t)}, \boldsymbol{a}_{t-1}, \boldsymbol{g}_{t}]
\end{equation}
comprises joint positions/velocities, base angular velocity, gravity projection, previous actions, and current commands. We design a factorized command space $\boldsymbol{g} = [\boldsymbol{g}_{loco}, \boldsymbol{g}_{torso}, \boldsymbol{g}_{arms}]^T$ enabling independent control of locomotion ($\boldsymbol{v}_{xy}, \omega_z, h_{pelvis}$), torso orientation (ZXY Euler angles), and arm joint targets (Table~\ref{tab:command_space}). Actions are target joint positions executed via PD control:
\begin{equation}
    \boldsymbol{a}_t = [\boldsymbol{q}_{legs}^{target}, \boldsymbol{q}_{torso}^{target}, \boldsymbol{q}_{arms}^{target}]^T \cdot \alpha_{scale} + \boldsymbol{q}_{default}
\end{equation}
where $\alpha_{scale} = 0.25$, and arm actions use residual modeling: $\boldsymbol{q}_{arms}^{final} = \boldsymbol{a}_{arms} + \boldsymbol{q}_{arms}^{desired}$ (Section~\ref{sec:residual_modeling}).

\subsection{Sequential Skill Acquisition and Adaptive Curriculum Learning}

To address the fundamental challenge of inefficient exploration in
high-dimensional command spaces, \ULC employs a \emph{sequential skill
    acquisition strategy} with adaptive command curriculum. The policy
progressively masters skills following a carefully designed hierarchical
sequence. This sequential approach prevents catastrophic forgetting and ensures
robust acquisition of fundamental capabilities before advancing to more complex
behaviors.

\subsubsection{Mathematical Framework for Curriculum Progression}

We formalize the curriculum learning process through a structured progression
system with rigorous mathematical foundations. Let $\mathcal{T} = \{T_1, T_2,
    T_3\}$ represent the ordered set of skills to be learned sequentially, where
\begin{align}
    T_1 & :\; \text{Base velocity tracking} \quad (\boldsymbol{v}_{xy}, \omega_z) \label{eq:skill_t1}                               \\
    T_2 & :\; \text{Base height tracking} \quad (h_{\text{pelvis}}) \label{eq:skill_t2}                                             \\
    T_3 & :\; \text{Torso and arm tracking} \quad (\boldsymbol{g}_{\text{torso}}, \boldsymbol{g}_{\text{arms}}) \label{eq:skill_t3}
\end{align}

For each skill $T_i$, we define a \emph{curriculum parameter} $\alpha_i(t) \in
    [0,1]$ that controls the difficulty progression over training time $t$. The
curriculum advancement follows a reward-based gating mechanism that evaluates
multiple performance metrics simultaneously.
\begin{equation}
    \alpha_i(t+1) = \begin{cases}
        \min\{1, \alpha_i(t) + \Delta\alpha\} & \text{if } \mathcal{C}_i(t) = \texttt{True} \\
        \alpha_i(t)                           & \text{otherwise}
    \end{cases}
    \label{eq:curriculum_update}
\end{equation}

where $\Delta\alpha = 0.05$ represents the curriculum increment. The
advancement conditions $\mathcal{C}_i(t)$ are specifically designed based on
empirical validation:

\begin{algorithm}[t]
    \caption{ULC Sequential Skill Acquisition with Adaptive Curriculum}
    \label{alg:ulc_curriculum}
    \begin{algorithmic}[1]
        \STATE \textbf{Input:} Skills $\mathcal{T} = \{T_1, T_2, T_3\}$, reward weights $\{w_{vel}, w_{height}, w_{upper}, w_{torso}, w_{hip}\}$
        \STATE \textbf{Initialize:} $\alpha_2 \leftarrow 0.0$, $\alpha_3 \leftarrow 0.0$, $t \leftarrow 0$
        \STATE \textbf{Initialize:} Active skills $\mathcal{A} \leftarrow \{T_1\}$, curriculum update interval $I \leftarrow 1000$ steps

        \WHILE{training not converged}
        \STATE $t \leftarrow t + 1$

        \STATE \textcolor{blue}{// Sample commands based on current curriculum}
        \FOR{each skill $T_i \in \mathcal{A}$}
        \STATE Sample commands $\boldsymbol{g}_i$ using curriculum parameter $\alpha_i$
        \ENDFOR

        \STATE \textcolor{blue}{// Execute training step}
        \STATE $\boldsymbol{g} \leftarrow$ Concatenate sampled commands from active skills
        \STATE Execute policy $\pi_{\boldsymbol{\theta}}(\boldsymbol{a}_t | \boldsymbol{s}_t, \boldsymbol{g})$
        \STATE Compute episode rewards and track running averages
        \STATE Update policy parameters $\boldsymbol{\theta}$ using PPO

        \STATE \textcolor{blue}{// Evaluate curriculum advancement every $I$ steps}
        \IF{$t \bmod I = 0$}
        \STATE \textcolor{blue}{// Height curriculum advancement}
        \IF{$\mathcal{C}_2(t)$ and $\alpha_2 < 0.98$}
        \STATE $\alpha_2 \leftarrow \min(0.98, \alpha_2 + 0.05)$
        \STATE Reset tracked rewards for next evaluation
        \ENDIF

        \STATE \textcolor{blue}{// Upper body curriculum advancement}
        \IF{$\mathcal{C}_3(t)$ and $\alpha_3 < 0.98$}
        \STATE $\alpha_3 \leftarrow \min(0.98, \alpha_3 + 0.05)$
        \STATE Reset tracked rewards for next evaluation
        \ENDIF

        \STATE \textcolor{blue}{// Activate terrain curriculum when both skills mastered}
        \IF{$\alpha_2 > 0.98$ and $\alpha_3 > 0.98$}
        \STATE Enable terrain level progression
        \ENDIF
        \ENDIF
        \ENDWHILE
    \end{algorithmic}
\end{algorithm}

\paragraph{Height Curriculum Advancement ($\mathcal{C}_2$)}

The height curriculum advancement condition implements a multi-criteria
evaluation that ensures the robot has mastered fundamental locomotion skills
before introducing height variation challenges. The condition is mathematically
defined as:
\begin{equation}
    \mathcal{C}_2(t) = \mathcal{C}_{\text{height}}(t) \land \mathcal{C}_{\text{velocity}}(t) \land \mathcal{C}_{\text{hip}}(t),
    \label{eq:height_curriculum}
\end{equation}
where each component evaluates specific performance metrics with carefully tuned thresholds:
\begin{align}
    \mathcal{C}_{\text{height}}(t)   & = R_{\text{height}}^{\text{avg}}(t) \geq \tau_{\text{height}} \label{eq:height_condition} \\
    \mathcal{C}_{\text{velocity}}(t) & = R_{\text{vel}}^{\text{avg}}(t) \geq \tau_{\text{vel}} \label{eq:velocity_condition}     \\
    \mathcal{C}_{\text{hip}}(t)      & = R_{\text{hip}}^{\text{avg}}(t) \geq \tau_{\text{hip}} \label{eq:hip_condition}
\end{align}

The threshold parameters are determined through systematic hyperparameter
tuning: $\tau_{\text{height}} = 0.85$ and $\tau_{\text{vel}} = 0.8$ ensure the
policy achieves near-convergent tracking performance before curriculum
advancement, while $\tau_{\text{hip}} = 0.2$ prevents excessive hip deviation
that would compromise subsequent skill learning. These values were selected
based on grid search to balance training stability with curriculum progression
speed.

\paragraph{Upper Body Curriculum Advancement ($\mathcal{C}_3$)}

The upper body curriculum advancement implements a comprehensive condition that
requires mastery of both arm tracking and torso control capabilities, while
simultaneously maintaining all previously acquired skills. The advancement
criterion is
\begin{equation}
    \mathcal{C}_3(t) = \mathcal{C}_{\text{upper}}(t) \land \mathcal{C}_{\text{torso}}(t) \land \mathcal{C}_{\text{prev}}(t) \land \mathcal{C}_{\text{complete}}(t)
    \label{eq:upper_curriculum}
\end{equation}

The individual components are rigorously defined as:
\begin{align}
    \mathcal{C}_{\text{upper}}(t)    & = R_{\text{upper}}^{\text{avg}}(t) \geq \tau_{\text{upper}} \label{eq:upper_condition}                                              \\
    \mathcal{C}_{\text{torso}}(t)    & = R_{\text{torso}}^{\text{avg}}(t) \geq \tau_{\text{torso}} \label{eq:torso_condition}                                              \\
    \mathcal{C}_{\text{prev}}(t)     & = \mathcal{C}_{\text{height}}(t) \land \mathcal{C}_{\text{velocity}}(t) \land \mathcal{C}_{\text{hip}}(t) \label{eq:prev_condition} \\
    \mathcal{C}_{\text{complete}}(t) & = \alpha_2 \geq 0.98 \label{eq:complete_condition}
\end{align}

where $R_{\text{upper}}^{\text{avg}}(t)$ denotes the upper body joint tracking
reward, and $R_{\text{torso}}^{\text{avg}}(t)$ represents the torso orientation
tracking reward, which are detailed in Appendix~\ref{subsec:append_reward}. The
threshold coefficients $\tau_{\text{upper}}=0.8$ and $\tau_{\text{torso}}=0.8$
are chosen to ensure adequate skill mastery; lower values led to premature
advancement and skill degradation, while higher values unnecessarily prolonged
training.

This multi-criteria gating mechanism ensures that curriculum progression occurs
only when all prerequisite skills are sufficiently mastered, thereby preventing
catastrophic forgetting and maintaining stable performance across all learned
capabilities.

\begin{figure*}[t]
    \centering
    \includegraphics[width=1.0\textwidth]{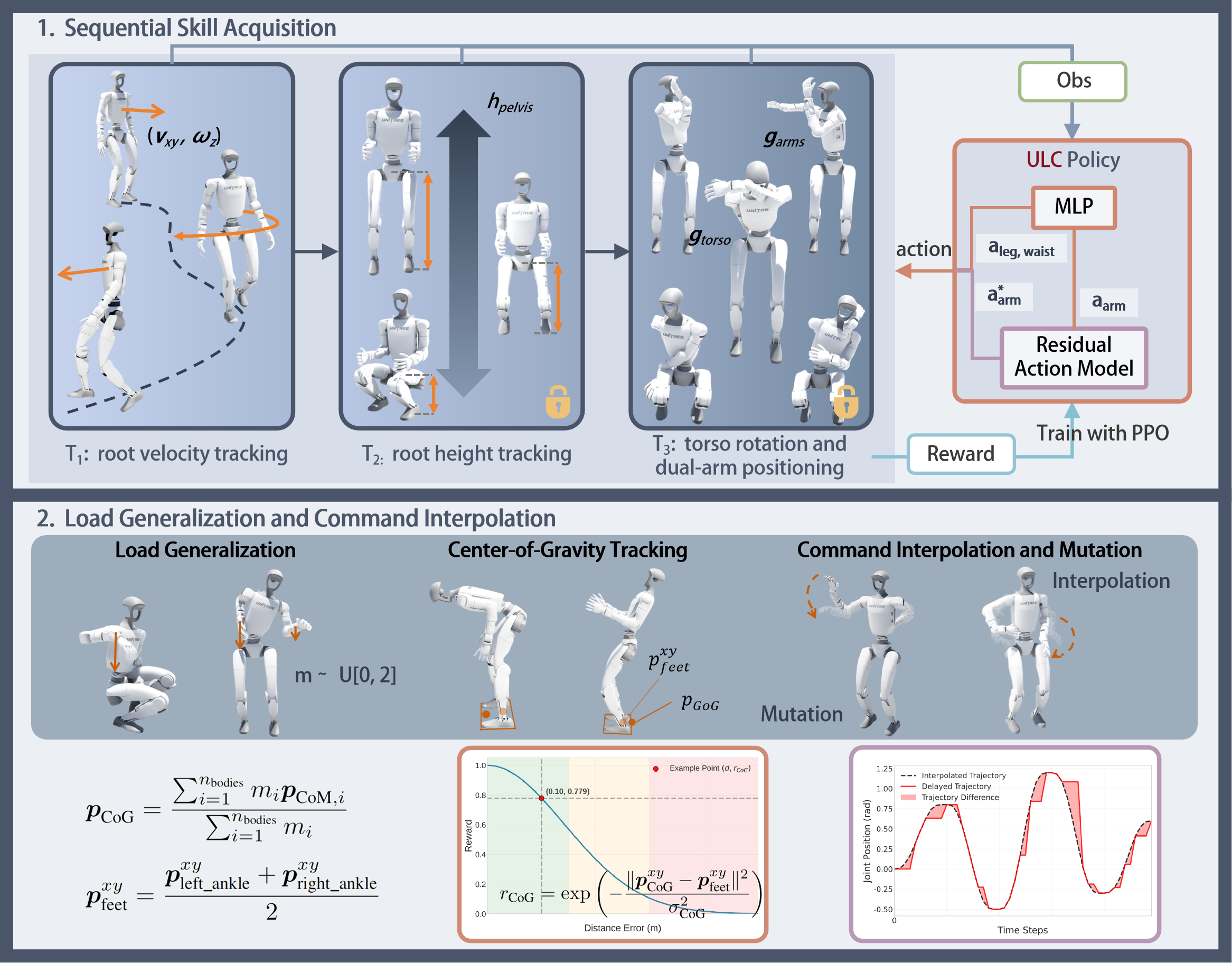}
    \caption{Method overview of the \textcolor{myred}{U}nified
        \textcolor{myred}{L}oco-Manipulation \textcolor{myred}{C}ontroller
        (\ULC). Our approach employs massively parallel reinforcement learning to train a single unified policy that tracks procedurally sampled commands including root velocity, root height, torso orientation, and arm joint positions. The framework addresses multi-task learning challenges through sequential skill acquisition with adaptive curriculum, deployment-realistic command generation with interpolation and random delay, and loaded balance optimization with center of mass tracking.}
    \label{fig:method}
    \vspace{-15pt}
\end{figure*}

\vspace{-10pt}
\subsection{Stochastic Delay Mechanism and Command Interpolation}
\label{sec:command_interpolation}
To ensure stable arm movements and enhance training robustness, we implement
sophisticated command processing mechanisms such as quintic polynomial
interpolation and stochastic delay modeling to accurately reflect real-world communication delays.

\paragraph{Quintic Polynomial Interpolation}

The upper body commands are smoothly interpolated using quintic polynomial
transitions between randomly sampled target positions. This design separates
trajectory smoothness from deployment realism: the quintic interpolation
generates physically plausible reference trajectories, while the subsequent
stochastic delay mechanism introduces realistic command discontinuities. The interpolation is
executed over a fixed interval $T_{\text{interval}} = 1.0$ s, with the
instantaneous target position determined by
\begin{equation}
    \boldsymbol{q}_{\text{target}}(t) = \boldsymbol{q}_{\text{start}} + (\boldsymbol{q}_{\text{goal}} - \boldsymbol{q}_{\text{start}}) \cdot s(t),
    \label{eq:quintic_interpolation}
\end{equation}
where $s(t)$ is the quintic smoothing factor
\begin{equation}
    s(t) = 10t^3 - 15t^4 + 6t^5, \quad t \in [0,1].
    \label{eq:quintic_smoothing}
\end{equation}

The movement step counter $t_{\text{step}}$ is normalized as $t =
    \min(t_{\text{step}} / T_{\text{interval}}, 1.0)$ to ensure smooth transitions.
This quintic polynomial ensures $C^2$ continuity with zero velocity and
acceleration at the endpoints, providing natural arm movement characteristics.


\paragraph{Stochastic Delay Mechanism}

The delay mechanism is implemented through a sophisticated accumulation and
release system that operates on the incremental commands between consecutive
timesteps. Let $\Delta \boldsymbol{q}^{(t)}$ represent the incremental change
in target position at timestep $t$
\begin{equation}
    \Delta \boldsymbol{q}^{(t)} = \boldsymbol{q}_{\text{target}}^{(t)} - \boldsymbol{q}_{\text{theoretical}}^{(t-1)},
\end{equation}
where $\boldsymbol{q}_{\text{theoretical}}^{(t-1)}$ is the theoretical position from
the previous timestep's interpolation.

\textbf{Delay Mask and Accumulation:} At each timestep, a random delay mask $\boldsymbol{d}^{(t)} \in \{0,1\}^{n_j}$ is generated
\begin{equation}
    d_j^{(t)} \sim \text{Bernoulli}(p_{delay}), \quad j = 1, \ldots, n_j,
\end{equation}
where $p_{delay} = 0.5$ is the fixed delay probability.

\textbf{Command Release:} The effective command executed at timestep $t$ releases both the current increment (if not delayed) and any previously accumulated commands (if released):
\begin{equation}
    \Delta \boldsymbol{q}_{\text{effective}}^{(t)} = \Delta \boldsymbol{q}^{(t)} \odot (1 - \boldsymbol{d}^{(t)}) + \boldsymbol{A}^{(t-1)} \odot (1 - \boldsymbol{d}^{(t)})
    \label{eq:command_release}
\end{equation}

\textbf{Buffer Update:} After command release, the accumulation buffer is updated to retain delayed commands and accumulate new delayed increments:
\begin{equation}
    \boldsymbol{A}^{(t)} = \boldsymbol{A}^{(t-1)} \odot \boldsymbol{d}^{(t)} + \Delta \boldsymbol{q}^{(t)} \odot \boldsymbol{d}^{(t)}
    \label{eq:buffer_update}
\end{equation}


\begingroup
\setlength{\tabcolsep}{3.5pt}
\begin{table*}[t]
    \centering
    \resizebox{1\linewidth}{!}{%
        \begin{tabular}{lc c ccccccc c ccccccc}
            \toprule
            \multirow{2}{*}{\textbf{Method}} &                          & \multicolumn{7}{c}{\textbf{Whole Command Space}} &                          & \multicolumn{7}{c}{\textbf{Edge Command Space}}                                                                                                      \\
            \cmidrule{3-9} \cmidrule{11-17}
                                             &                          & $E_{v}\downarrow$                                & $E_{\omega}\downarrow$   & $E_{h}\downarrow$                               & $E_{y}\downarrow$        & $E_{p}\downarrow$        & $E_{r}\downarrow$        & $E_{a}\downarrow$
                                             &                          & $E_{v}\downarrow$                                & $E_{\omega}\downarrow$   & $E_{h}\downarrow$                               & $E_{y}\downarrow$        & $E_{p}\downarrow$        & $E_{r}\downarrow$        & $E_{a}\downarrow$ \\
            \midrule
            HOMIE                            &
                                             & 0.15\ci{.02}             & 0.18\ci{.02}                                     & 0.04\ci{.01}             & 0.08\ci{.01}                                    & /                        & /                        & 0.12\ci{.02}             &
                                             & 0.18\ci{.03}             & 0.22\ci{.03}                                     & 0.06\ci{.01}             & 0.12\ci{.02}                                    & /                        & /                        & 0.15\ci{.02}                                 \\
            HOMIE-3-DoF-Waist                       &
                                             & 0.14\ci{.02}             & 0.17\ci{.02}                                     & 0.04\ci{.01}             & 0.09\ci{.01}                                    & 0.15\ci{.02}             & 0.14\ci{.02}             & 0.12\ci{.02}             &
                                             & 0.25\ci{.04}             & 0.28\ci{.04}                                     & 0.10\ci{.02}             & 0.18\ci{.03}                                    & 0.28\ci{.04}             & 0.26\ci{.04}             & 0.18\ci{.03}                                 \\
            FALCON                           &
                                             & 0.16\ci{.02}             & 0.19\ci{.02}                                     & 0.05\ci{.01}             & 0.10\ci{.01}                                    & /                        & /                        & \underline{0.08\ci{.01}} &
                                             & 0.24\ci{.03}             & 0.26\ci{.03}                                     & 0.09\ci{.02}             & 0.16\ci{.02}                                    & /                        & /                        & 0.14\ci{.02}                                 \\
            AMO                              &
                                             & \textbf{0.08\ci{.01}}    & \textbf{0.10\ci{.01}}                            & \underline{0.03\ci{.01}} & \underline{0.07\ci{.01}}                        & 0.11\ci{.02}             & 0.12\ci{.02}             & 0.11\ci{.02}             &
                                             & \underline{0.12\ci{.02}} & \underline{0.14\ci{.02}}                         & \underline{0.05\ci{.01}} & \underline{0.10\ci{.02}}                        & \underline{0.16\ci{.02}} & \underline{0.15\ci{.02}} & 0.14\ci{.02}                                 \\
            R$^2$S$^2$                       &
                                             & 0.13\ci{.02}             & 0.15\ci{.02}                                     & 0.04\ci{.01}             & 0.17\ci{.02}                                    & 0.13\ci{.02}             & /                        & 0.10\ci{.01}             &
                                             & 0.17\ci{.02}             & 0.20\ci{.03}                                     & 0.07\ci{.01}             & 0.18\ci{.02}                                    & 0.19\ci{.03}             & /                        & \underline{0.13\ci{.02}}                     \\
            \textbf{ULC}                     &
                                             & \underline{0.10\ci{.01}} & \underline{0.12\ci{.01}}                         & \textbf{0.02\ci{.00}}    & \textbf{0.05\ci{.01}}                           & \textbf{0.06\ci{.01}}    & \textbf{0.05\ci{.01}}    & \textbf{0.04\ci{.01}}    &
                                             & \textbf{0.11\ci{.01}}    & \textbf{0.13\ci{.02}}                            & \textbf{0.03\ci{.01}}    & \textbf{0.06\ci{.01}}                           & \textbf{0.08\ci{.01}}    & \textbf{0.07\ci{.01}}    & \textbf{0.05\ci{.01}}                        \\
            \midrule
            \multirow{2}{*}{\textbf{Method}} &                          & \multicolumn{7}{c}{\textbf{Wrist Loaded (2kg)}}  &                          & \multicolumn{7}{c}{\textbf{Command Mutation}}                                                                                                        \\
            \cmidrule{3-9} \cmidrule{11-17}
                                             &                          & $E_{v}\downarrow$                                & $E_{\omega}\downarrow$   & $E_{h}\downarrow$                               & $E_{y}\downarrow$        & $E_{p}\downarrow$        & $E_{r}\downarrow$        & $E_{a}\downarrow$
                                             &                          & $E_{v}\downarrow$                                & $E_{\omega}\downarrow$   & $E_{h}\downarrow$                               & $E_{y}\downarrow$        & $E_{p}\downarrow$        & $E_{r}\downarrow$        & $E_{a}\downarrow$ \\
            \midrule
            HOMIE                            &
                                             & 0.18\ci{.02}             & 0.21\ci{.03}                                     & 0.05\ci{.01}             & 0.10\ci{.01}                                    & /                        & /                        & 0.18\ci{.03}             &
                                             & 0.28\ci{.04}             & 0.32\ci{.05}                                     & 0.12\ci{.02}             & 0.20\ci{.03}                                    & /                        & /                        & 0.22\ci{.03}                                 \\
            HOMIE-3-DoF-Waist                       &
                                             & 0.17\ci{.02}             & 0.20\ci{.03}                                     & 0.05\ci{.01}             & 0.11\ci{.02}                                    & 0.18\ci{.03}             & 0.17\ci{.02}             & 0.17\ci{.02}             &
                                             & 0.30\ci{.05}             & 0.35\ci{.05}                                     & 0.14\ci{.03}             & 0.24\ci{.04}                                    & 0.32\ci{.05}             & 0.30\ci{.04}             & 0.24\ci{.04}                                 \\
            FALCON                           &
                                             & 0.18\ci{.02}             & 0.21\ci{.03}                                     & 0.06\ci{.01}             & 0.12\ci{.02}                                    & /                        & /                        & \underline{0.11\ci{.02}} &
                                             & 0.26\ci{.04}             & 0.29\ci{.04}                                     & 0.10\ci{.02}             & 0.18\ci{.03}                                    & /                        & /                        & \underline{0.16\ci{.02}}                     \\
            AMO                              &
                                             & \textbf{0.09\ci{.01}}    & \textbf{0.11\ci{.01}}                            & \underline{0.04\ci{.01}} & \underline{0.08\ci{.01}}                        & 0.13\ci{.02}             & 0.14\ci{.02}             & 0.16\ci{.02}             &
                                             & \underline{0.14\ci{.02}} & \underline{0.16\ci{.02}}                         & \underline{0.06\ci{.01}} & 0.14\ci{.02}                                    & 0.22\ci{.03}             & 0.20\ci{.03}             & 0.18\ci{.03}                                 \\
            R$^2$S$^2$                       &
                                             & 0.15\ci{.02}             & 0.17\ci{.02}                                     & 0.05\ci{.01}             & 0.19\ci{.02}                                    & 0.15\ci{.02}             & /                        & 0.14\ci{.02}             &
                                             & 0.20\ci{.03}             & 0.23\ci{.03}                                     & 0.08\ci{.01}             & 0.22\ci{.04}                                    & 0.20\ci{.03}             & /                        & 0.17\ci{.02}                                 \\
            \textbf{ULC}                     &
                                             & \underline{0.10\ci{.01}} & \underline{0.13\ci{.02}}                         & \textbf{0.03\ci{.01}}    & \textbf{0.06\ci{.01}}                           & \textbf{0.07\ci{.01}}    & \textbf{0.06\ci{.01}}    & \textbf{0.05\ci{.01}}    &
                                             & \textbf{0.12\ci{.02}}    & \textbf{0.14\ci{.02}}                            & \textbf{0.03\ci{.01}}    & \textbf{0.07\ci{.01}}                           & \textbf{0.09\ci{.01}}    & \textbf{0.08\ci{.01}}    & \textbf{0.06\ci{.01}}                        \\
            \bottomrule
        \end{tabular}}
    \caption{\textbf{Tracking accuracy comparison across different scenarios.} We present a performance comparison between \ULC and baselines for the proposed metrics. The means and standard deviation are reported across 5 evaluations, each with 1024 parallel environments for 50,000 steps. Best results are in \textbf{bold}, second best are \underline{underlined}. ``/'' indicates the method lacks this capability.}
    \label{tab:tracking_accuracy}
    \vspace{-0.4cm}
\end{table*}
\endgroup

\vspace{-10pt}
\subsection{Load Generalization and Balance Control}


\paragraph{Random Load Distribution}

During training, we apply random masses to the robot's wrists to simulate
diverse payload conditions. The mass randomization is applied to the robot's
wrists masses during environment reset, with the total wrists mass distribution
modified to simulate carrying loads.

\paragraph{Center of Mass Tracking}

We implement a sophisticated center of mass tracking reward that maintains
stability across all motion phases. The reward function is formulated as:

\begin{equation}
    r_{\text{CoM}} = \exp\left(-\frac{\|\boldsymbol{p}_{\text{CoM}}^{xy} - \boldsymbol{p}_{\text{feet}}^{xy}\|^2}{\sigma_{\text{CoM}}^2}\right)
    \label{eq:cog_reward}
\end{equation}

where $\boldsymbol{p}_{\text{CoM}}^{xy}$ is the horizontal projection of the
whole-body center of mass, and $\boldsymbol{p}_{\text{feet}}^{xy}$
represents the midpoint between the ankle positions.



\textbf{Feet Support Reference:} The support reference is computed as the midpoint between the ankle positions

\begin{equation}
    \boldsymbol{p}_{\text{feet}}^{xy} = \frac{\boldsymbol{p}_{\text{left\_ankle}}^{xy} + \boldsymbol{p}_{\text{right\_ankle}}^{xy}}{2}.
    \label{eq:feet_reference}
\end{equation}

This provides a consistent reference point for balance control that accounts
for the robot's current stance configuration.


\subsection{Residual Action Modeling for Arm Control}
\label{sec:residual_modeling}

We introduce residual action modeling for arm joints that enables precise
tracking while maintaining training stability. The final control command
combines policy output with residual correction:
\begin{equation}
    \boldsymbol{q}_{\text{processed}} = \alpha_{scale} \cdot \pi_{\boldsymbol{\theta}}(\boldsymbol{s}, \boldsymbol{g}) + \boldsymbol{q}_{\text{default}}
    \label{eq:policy_processing}
\end{equation}
\vspace{-25pt}

\begin{equation}
    \boldsymbol{q}_{\text{final}}[\mathcal{J}_{\text{upper}}] = \boldsymbol{q}_{\text{processed}}[\mathcal{J}_{\text{upper}}] + \boldsymbol{q}_{\text{desired}}[\mathcal{J}_{\text{upper}}],
    \label{eq:residual_application}
\end{equation}
where $\boldsymbol{q}_{\text{desired}}$ is generated through command interpolation and delay mechanism. The residual term acts as a feedforward component compensating for predictable dynamics, allowing the policy to focus on learning corrective adjustments rather than reconstructing the entire control signal.




\section{EXPERIMENT}
\label{sec:exp}
\begin{figure*}[htbp]
    \vspace{5pt}
    \centering
    \includegraphics[width=1.0 \linewidth]{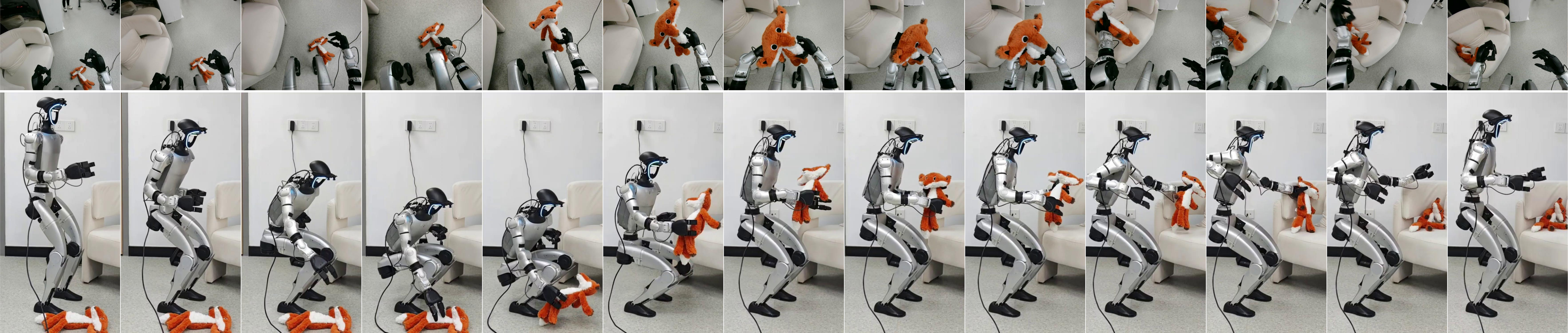}
    \vspace{5pt}
    \centering
    \includegraphics[width=1.0 \linewidth]{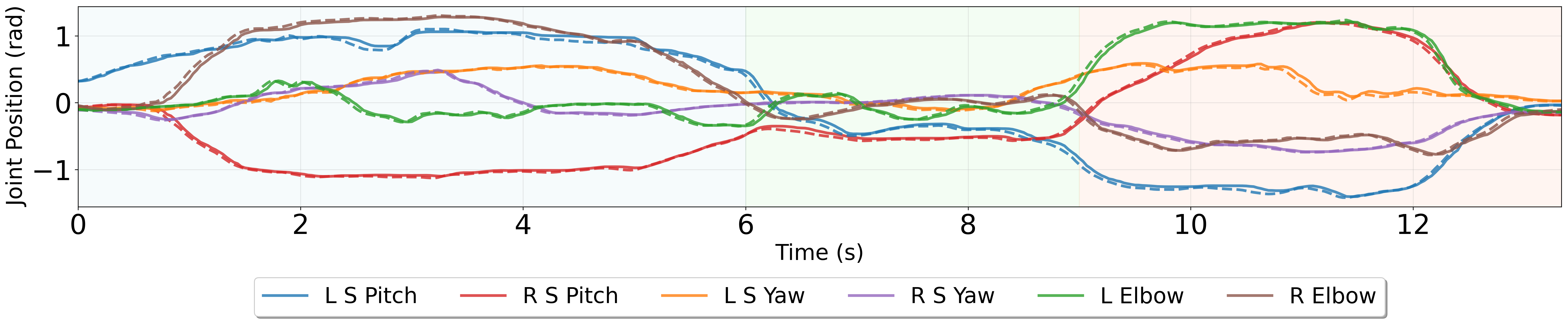}
    \vspace{-20pt}
    \caption{Time-series visualization of the doll pick-and-place task, covering all key stages: squatting to pick up the doll, hand switching, and placing the doll at the target location.}
    \label{fig:doll_pick}
\end{figure*}

\subsection{Experimental Setup}

We compare \ULC with state-of-the-art loco-manipulation controllers. Baseline
results are obtained from official checkpoints or our faithful
reimplementations. All methods use identical simulator settings, evaluation
protocols, and low-level assumptions (e.g., PD gains, control frequency).
Methods with different observation/action interfaces (e.g., MoCap-driven arms,
skill primitives) are evaluated in their native interfaces within feasible
command ranges. For robustness evaluation, identical payload and command
mutation conditions are applied, while training-time randomization follows each
method's original design.

\begin{itemize}

    \item \textbf{HOMIE}~\cite{benHOMIEHumanoidLocoManipulation2025}: A decoupled
          controller using RL for legs and PD control for waist yaw and arms.

    \item \textbf{HOMIE-3-DoF-Waist}: An extension of HOMIE with unlocked three waist DoF for PD control.

    \item \textbf{FALCON}~\cite{zhangFALCONLearningForceAdaptive2025}: A decoupled
          controller using dual policy for lower and upper body with adaptive force curriculum.

    \item \textbf{AMO}~\cite{liAMOAdaptiveMotion2025}: A hierarchical controller
          combining trajectory optimization with RL for leg and waist control, with PD-controlled arms.

    \item \textbf{R$^2$S$^2$}~\cite{zhangUnleashingHumanoidReaching2025}: A skill-based whole-body controller using a pre-trained skill library for goal-reaching tasks.

\end{itemize}

Our metrics include:
\begin{itemize}
    \item \textbf{Root Linear Velocity Tracking Error} $E_{v}$
    \item \textbf{Root Angular Velocity Tracking Error} $E_{\omega}$
    \item \textbf{Root Height Tracking Error} $E_{h}$
    \item \textbf{Root Yaw Orientation Tracking Error} $E_{y}$
    \item \textbf{Root Pitch Orientation Tracking Error} $E_{p}$
    \item \textbf{Root Roll Orientation Tracking Error} $E_{r}$
    \item \textbf{Arm Joint Position Tracking Error} $E_{a}$
\end{itemize}

All metrics are computed by rolling out 1024 parallel environments in
Isaaclab~\cite{Mittal_2023} for 50,000 steps, averaging tracking errors across
all timesteps and environments.

\begin{table}[t]
    \centering

    \resizebox{\columnwidth}{!}{%
        \begin{tabular}{l|cc|cc|cc|cc|c}
            \toprule
                              & \multicolumn{2}{c|}{\textbf{Height}} & \multicolumn{2}{c|}{\textbf{Yaw}} & \multicolumn{2}{c|}{\textbf{Pitch}} & \multicolumn{2}{c|}{\textbf{Roll}} & \textbf{Arm}                                                                     \\
            \cmidrule{2-9}
            \textbf{Method}   & Min                                  & Max                               & Min                                 & Max                                & Min            & Max           & Min            & Max           & \textbf{Ctrl}  \\
            \midrule
            HOMIE             & 0.30                                 & 0.75                              & -2.62                               & 2.62                               & /              & /             & /              & /             & PD             \\
            HOMIE-3-DoF-Waist & 0.30                                 & 0.75                              & -2.62                               & 2.62                               & -0.52          & 0.52          & -0.52          & 0.52          & PD             \\
            FALCON            & 0.50                                 & 0.75                              & -1.00                               & 1.00                               & /              & /             & /              & /             & MoCap          \\
            AMO               & 0.35                                 & 0.75                              & -2.62                               & 2.62                               & -0.52          & 1.57          & -0.46          & 0.46          & PD             \\
            R$^2$S$^2$        & 0.35                                 & 0.75                              & -1.00                               & 1.00                               & 0.00           & 0.50          & /              & /             & Skill Lib      \\
            \midrule
            \textbf{ULC}      & \textbf{0.30}                        & \textbf{0.75}                     & \textbf{-2.62}                      & \textbf{2.62}                      & \textbf{-0.52} & \textbf{1.57} & \textbf{-0.52} & \textbf{0.52} & \textbf{Proc.} \\
            \bottomrule
        \end{tabular}%
    }
    \caption{Reachable command ranges for different methods.}
    \label{tab:tracking_ranges}
    \vspace{-10pt}
\end{table}

\subsection{Comparison of Reachable Workspace}

Table~\ref{tab:tracking_ranges} compares the reachable command ranges across
different methods.

\textbf{HOMIE} and \textbf{HOMIE-3-DoF-Waist} employ PD control for waist joints, creating a decoupling between torso rotation and leg control. Although this enables full yaw rotation and maximum root height range, the legs cannot actively participate in torso orientation control. \textbf{FALCON} prioritizes adaptive force curriculum learning but suffers from restricted yaw range and elevated minimum root height, with dual-arm control relying entirely on motion capture data. \textbf{AMO} addresses the out-of-distribution (OOD) limitation through its motion adaptation module, achieving asymmetric pitch control, but remains constrained in roll orientation. \textbf{R$^2$S$^2$} utilizes a pre-defined skill library, but the reliance on pre-defined primitives constrains torso rotation capabilities.

\ULC overcomes these limitations through unified coordinated control. Our approach achieves the maximum root height range, complete torso rotation tracking across all axes, and procedurally sampled dual-arm control without MoCap constraints.

\subsection{Comparison of Tracking Accuracy}

Table \ref{tab:tracking_accuracy} evaluates tracking performance across four
scenarios with commands sampled within each method's operational ranges: (1)
\textbf{Whole command space}; (2) \textbf{Edge command space}: extreme torso
rotation cases; (3) \textbf{Wrist loaded}: 2kg external loads; (4)
\textbf{Command mutation}: random command
delays~(\ref{sec:command_interpolation}).

\textbf{Locomotion Control:} AMO demonstrates exceptional linear and angular velocity tracking due to its hierarchical design combining trajectory optimization with RL. AMO and \ULC achieves competitive performance through unified whole-body control, while HOMIE and FALCON show moderate performance due to decoupled architectures.

\textbf{Torso Orientation Control:} \ULC, HOMIE-3-DoF-Waist, and AMO support full 3-DoF torso control. \ULC excels in yaw tracking and achieves superior pitch and roll control. AMO shows competitive yaw but higher pitch and roll errors. HOMIE-3-DoF-Waist relies entirely on PD control, resulting in degraded accuracy. HOMIE and FALCON lack pitch/roll capabilities, while R$^2$S$^2$ provides pitch and yaw control but lacks roll control capability.

\textbf{Dual-Arm Tracking:} HOMIE and AMO rely on PD controllers, achieving moderate performance. FALCON's upper-body RL policy achieves better tracking but remains constrained by MoCap dependency. \ULC outperforms all methods through residual action modeling and sequential skill acquisition.

\textbf{Robustness Under Extreme Conditions:} In edge command space scenarios, HOMIE-3-DoF-Waist suffers severe degradation due to inadequate coordination between PD-controlled torso and RL-controlled legs. \ULC maintains robust performance owing to its unified architecture.

\textbf{External Load Adaptation:} Under 2kg wrist loads, AMO maintains its locomotion advantage, but PD-based arm control shows noticeable degradation. FALCON's force-adaptive curriculum provides partial load robustness, while \ULC achieves superior load adaptation across all metrics.

\textbf{Command Mutation Robustness:} Under stochastic command delays, \ULC demonstrates superior robustness, while other methods show significant deterioration. AMO experiences substantial degradation in torso control, and HOMIE variants suffer severe performance loss.

\begin{figure*}[htbp]
    \vspace{5pt}
    \centering
    \includegraphics[width=1.0 \linewidth]{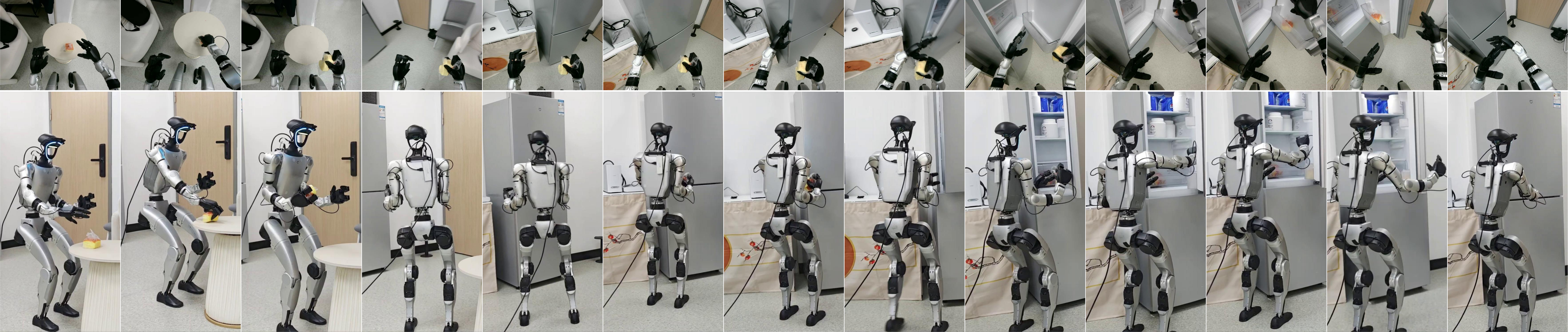}
    \centering
    \includegraphics[width=1.0 \linewidth]{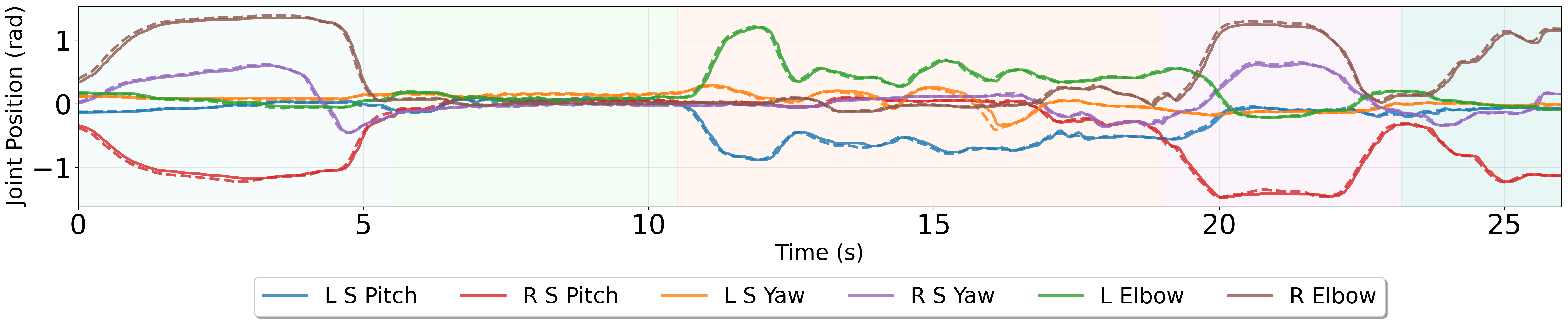}
    \caption{Time-series visualization of the refrigerator task, covering all five stages: picking up the bread, walking to the refrigerator, opening the door, placing the bread inside, and closing the door.}
    \label{fig:refrigerator_put}
    \vspace{-10pt}
\end{figure*}

\subsection{Ablation on Policy Training}

We evaluate the contribution of each key component by removing one module at a
time: Sequential Skill Acquisition, Residual Action Model, Load Randomization,
and center of mass Tracking. The results are visualized in
Table~\ref{tab:ablation}, with evaluation performed under 2kg wrist load.

Removing Sequential Skill Acquisition leads to significant degradation in torso
orientation control ($E_p$: +100\%, $E_r$: +100\%), confirming the importance
of progressive skill composition. Excluding the Residual Action Model causes
the largest arm tracking degradation ($E_a$: +140\%), validating its role in
fine-grained upper body control. Without Load Randomization, locomotion and arm
tracking deteriorate notably under payload. Omitting center of mass Tracking
results in the most severe balance degradation ($E_v$: +10\%, $E_\omega$:
+38\%, $E_h$: +100\%), demonstrating its essential role in dynamic stability.

The full \ULC model achieves the lowest errors across all metrics, validating
that each component is indispensable for robust, high-precision
loco-manipulation.

\begin{table}[h]
    \centering
    \vspace{2pt}
    \setlength{\tabcolsep}{4pt}
    \resizebox{\columnwidth}{!}{%
        \begin{tabular}{l|c|c|c|c|c|c|c}
            \toprule
            \textbf{Method}     & \makecell{$E_v$                                                                                                                                                                                                                                                                                                                                                                                                                                                                                                              \\(m/s)} & \makecell{$E_\omega$\\(rad/s)} & \makecell{$E_h$\\(m)} & \makecell{$E_y$\\(rad)} & \makecell{$E_p$\\(rad)} & \makecell{$E_r$\\(rad)} & \makecell{$E_a$\\(rad)} \\
            \midrule
            w/o Seq. Skill Acq. & \makecell{0.11\\{\scriptsize$\pm$0.02}}          & \makecell{0.17\\{\scriptsize$\pm$0.02}}          & \makecell{0.05\\{\scriptsize$\pm$0.01}}          & \makecell{0.10\\{\scriptsize$\pm$0.02}}          & \makecell{0.14\\{\scriptsize$\pm$0.02}}          & \makecell{0.12\\{\scriptsize$\pm$0.02}}          & \makecell{0.09\\{\scriptsize$\pm$0.02}}          \\
            \midrule
            w/o Residual Action & \makecell{0.12\\{\scriptsize$\pm$0.01}}          & \makecell{0.15\\{\scriptsize$\pm$0.02}}          & \makecell{0.04\\{\scriptsize$\pm$0.01}}          & \makecell{0.08\\{\scriptsize$\pm$0.01}}          & \makecell{0.10\\{\scriptsize$\pm$0.02}}          & \makecell{0.09\\{\scriptsize$\pm$0.01}}          & \makecell{0.12\\{\scriptsize$\pm$0.02}}          \\
            \midrule
            w/o Load Random.    & \makecell{0.11\\{\scriptsize$\pm$0.02}}          & \makecell{0.16\\{\scriptsize$\pm$0.02}}          & \makecell{0.04\\{\scriptsize$\pm$0.01}}          & \makecell{0.08\\{\scriptsize$\pm$0.01}}          & \makecell{0.09\\{\scriptsize$\pm$0.01}}          & \makecell{0.08\\{\scriptsize$\pm$0.01}}          & \makecell{0.10\\{\scriptsize$\pm$0.02}}          \\
            \midrule
            w/o CoM Tracking    & \makecell{0.11\\{\scriptsize$\pm$0.02}}          & \makecell{0.18\\{\scriptsize$\pm$0.03}}          & \makecell{0.06\\{\scriptsize$\pm$0.01}}          & \makecell{0.09\\{\scriptsize$\pm$0.02}}          & \makecell{0.11\\{\scriptsize$\pm$0.02}}          & \makecell{0.10\\{\scriptsize$\pm$0.02}}          & \makecell{0.08\\{\scriptsize$\pm$0.01}}          \\
            \midrule
            \textbf{ULC (Ours)} & \makecell{\textbf{0.10}\\{\scriptsize$\pm$0.01}} & \makecell{\textbf{0.13}\\{\scriptsize$\pm$0.02}} & \makecell{\textbf{0.03}\\{\scriptsize$\pm$0.01}} & \makecell{\textbf{0.06}\\{\scriptsize$\pm$0.01}} & \makecell{\textbf{0.07}\\{\scriptsize$\pm$0.01}} & \makecell{\textbf{0.06}\\{\scriptsize$\pm$0.01}} & \makecell{\textbf{0.05}\\{\scriptsize$\pm$0.01}} \\
            \bottomrule
        \end{tabular}%
    }
    \caption{Ablation study on policy training components. Best results are in \textbf{bold}.}
    \label{tab:ablation}
    \vspace{-10pt}
\end{table}

\vspace{-10pt}
\subsection{Real World Results}

We evaluate \ULC in real-world scenarios to validate how its height control,
torso rotation capabilities, and dual-arm tracking precision contribute to
practical task performance.

\subsubsection{\textbf{Teleoperation Results}}

We evaluate two representative teleoperation scenarios requiring coordinated
locomotion and manipulation, as illustrated in Fig.~\ref{fig:doll_pick} and
Fig.~\ref{fig:refrigerator_put}.

\textbf{Pick and place the doll on the sofa}: This task evaluates \ULC's ability to perform coordinated whole-body manipulation: (1) squatting down and grasping the doll using precise height control and torso pitch adjustment; (2) standing up and passing the doll to the other hand; (3) placing the doll onto the sofa. As shown in Fig.~\ref{fig:doll_pick}, the execution demonstrates \ULC's locomotion stability during height transitions and dynamic balance during coordinated arm movements. The tracking curves show consistently low dual-arm tracking error throughout execution.

\textbf{Put the bread in the refrigerator}: This task demonstrates \ULC's ability to execute a complex, multi-step sequence: (1) grasp the bread from the table; (2) walk to the refrigerator while maintaining a secure hold; (3) open the refrigerator door with the left hand; (4) place the bread inside with accurate positioning; (5) close the door after releasing the bread. Fig.~\ref{fig:refrigerator_put} presents the full teleoperated process, with consistently low arm tracking error highlighting \ULC's precision in practical force-interactive loco-manipulation scenarios.

\section{CONCLUSIONS and LIMITATIONS}
\label{sec:conclusion}
We presented \textcolor{myred}{ULC}, a unified controller for humanoid
loco-manipulation that simultaneously achieve unified whole-body control, large
operational workspace, and high-precision tracking. By integrating all degrees
of freedom in a single controller with principled procedural command sampling,
\textcolor{myred}{ULC} enables robust and versatile performance across diverse
tasks and challenging scenarios. Extensive experiments demonstrate that
\textcolor{myred}{ULC} outperforms prior decoupled or MoCap-based methods in
tracking accuracy, workspace coverage, and robustness. Ablation studies further
confirm the necessity of each component in our framework.

A current limitation is that simplified locomotion commands preclude complex
leg patterns achievable through motion capture approaches. Addressing this
limitation to enhance locomotion expressiveness for more complex real-world
tasks will be the focus of our future work.

\clearpage
\printbibliography

\clearpage
\appendix
\section{APPENDIX}
\label{sec:appendix}
\subsection{Communication Architecture}
\label{subsec:append_communication}

The teleoperation system employs a distributed communication architecture that
coordinates multiple components across different computational nodes. Dual
cameras mounted on the robot's onboard computer transmit stereo images to the
host computer via TCP and ZeroMQ protocols. The host computer processes these
images for VR visualization while receiving operator commands through a network
router connection. Robot actuators (dexterous hands and joints) communicate
bidirectionally with the host computer using DDS protocol, transmitting states
and receiving action commands. The system uses an asynchronous architecture
where the teleoperation solver processes VR inputs to generate robot commands,
while a separate deployment module runs at 50Hz to continuously read and
execute the latest commands. This design ensures responsive control while
maintaining system modularity.

The complete communication architecture can be summarized as follows:

\begin{align}
    \text{Cameras}    & \xrightarrow{\text{TCP/ZeroMQ}} \text{Host} \xrightarrow{\text{Network}} \text{VR Headset} \label{eq:image_flow}   \\
    \text{VR Headset} & \xrightarrow{\text{Network}} \text{Host} \xrightarrow{\text{DDS}} \text{Solver} \label{eq:command_flow}            \\
    \text{Solver}     & \xrightarrow{\text{DDS}} \text{Deployment} \xrightarrow{\text{DDS}} \text{Robot Actuators} \label{eq:control_flow}
\end{align}

This distributed architecture enables scalable and responsive teleoperation
while maintaining the modularity necessary for system development and
debugging.


\subsection{Domain Randomization}
\label{subsec:append_dr}
We use domain randomization to simulate the sensor noise and physical variations in the real-world. The randomization parameters are shown in Table~\ref{tab:domain_random}.

\begin{table}[h]
    \centering
    \small
    \begin{tabular}{l c c c}
        \toprule
        \textbf{Parameter} & \textbf{Unit} & \textbf{Range} & \textbf{Operator} \\
        \midrule
        Angular Velocity   & rad/s         & $\pm 0.2$      & additive           \\
        Projected Gravity  & -             & $\pm 0.05$     & additive           \\
        Joint Position     & rad           & $\pm 0.01$     & additive           \\
        Joint Velocity     & rad/s         & $\pm 1.5$      & additive           \\
        Static Friction    & -             & [0.7, 1.0]     & uniform           \\
        Dynamic Friction   & -             & [0.4, 0.7]     & uniform           \\
        Restitution        & -             & [0.0, 0.005]   & uniform           \\
        Wrist Mass         & kg            & [0.0, 2.0]     & additive          \\
        Base Mass          & kg            & [-5.0, 5.0]    & additive          \\
        \bottomrule
    \end{tabular}
    \caption{Domain randomization parameters. Additive randomization adds a random value within a specified range to the parameter, while scaling randomization adjusts the parameter by a random multiplication factor within the range.}
    \label{tab:domain_random}
\end{table}

\begin{figure*}[htbp]
    \vspace{5pt}
    \centering
    \includegraphics[width=1.0 \linewidth]{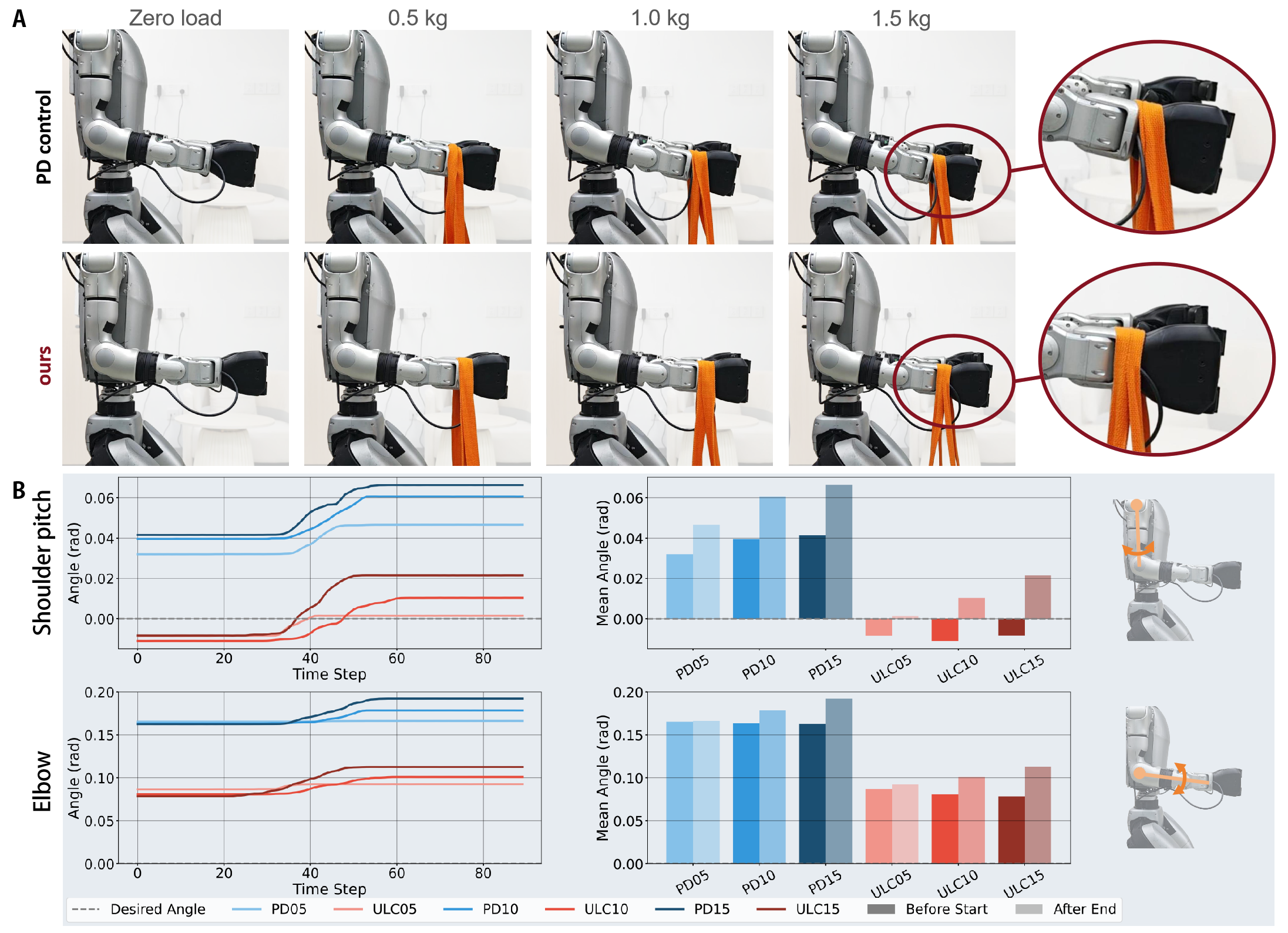}
    \caption{Comparison of joint angle tracking errors for \ULC and traditional PD control (gains: $K_p=80$, $K_d=3$) under different external loads (0.5~kg, 1.0~kg, 1.5~kg) in real-world experiments. \ULC consistently achieves lower errors than PD control at all load levels, demonstrating superior force adaptation and robustness to external disturbances.}
    \label{fig:load_comparison}
    \vspace{-5pt}
\end{figure*}

\subsection{Real world Loaded Comparison}
We conduct a controlled experiment comparing \ULC with traditional PD control
under external wrist loads of 0.5~kg, 1.0~kg, and 1.5~kg. \ULC and PD
controller share the same PD parameters (PD gains: $K_p=80$, $K_d=3$). Both
methods are required to maintain dual-arm poses with target joint angles set to
zero (forearms parallel to the ground). The tracking errors under each load
condition are visualized in Fig.~\ref{fig:load_comparison}.

Across all load levels, \ULC consistently achieves lower joint angle deviations
than PD control. Notably, even at the highest load of 1.5~kg, \ULC maintains
high tracking accuracy, while PD control exhibits significant errors due to
inadequate gravity compensation. This performance gap is evident at every
tested load (0.5~kg, 1.0~kg, 1.5~kg), where \ULC's learned dynamics naturally
incorporate force adaptation, resulting in superior robustness and precision.
In contrast, PD control struggles to maintain parallel positioning even without
load, and its errors increase substantially as the load increases. These
results validate the advantage of \ULC in real-world manipulation tasks
requiring reliable force adaptation and precise tracking under varying external
disturbances.


\subsection{Reward Function}
\label{subsec:append_reward}
Our reward function is a sum of the following terms:

\begin{itemize}

    \item \textbf{Tracking Linear Velocity Reward} ($r_{\text{vel}}$): This term encourages the robot to track the commanded linear velocity in the $\mathit{xy}$-plane.
          \begin{equation*}
              r_{vel} := \exp(-\|v_{xy} - v_{xy}^*\|_2^2/\sigma_{vel}^2),
          \end{equation*}
          where $v_{xy}$ and $v_{xy}^*$ represent the actual and commanded linear velocities, respectively. $\sigma_{vel}$ is set to 0.5. Weight: 1.0.

    \item \textbf{Tracking Angular Velocity Reward} ($r_{\text{ang}}$): This term encourages the robot to track the commanded angular velocity.
          \begin{equation*}
              r_{ang} := \exp(-\|\omega_z - \omega_z^*\|_2^2/\sigma_{ang}^2),
          \end{equation*}
          where $\omega_z$ and $\omega_z^*$ represent the actual and commanded angular velocities, respectively. $\sigma_{ang}$ is set to 0.5. Weight: 1.25.

    \item \textbf{Root Height Tracking Reward} ($r_{\text{height}}$): This term encourages tracking of the commanded pelvis height.
          \begin{equation*}
              r_{height} := \exp(-|h - h^*|^2/\sigma_{height}^2),
          \end{equation*}
          where $h$ and $h^*$ are the actual and commanded root heights. $\sigma_{height}$ is set to 0.4. Weight: 1.0.

    \item \textbf{Upper Body Position Tracking Reward} ($r_{\text{upper}}$): This term encourages tracking of arm joint positions.
          \begin{equation*}
              r_{upper} := \exp(-\|\boldsymbol{q}_{upper} - \boldsymbol{q}_{upper}^*\|_2^2/\sigma_{upper}^2),
          \end{equation*}
          where $\boldsymbol{q}_{upper}$ and $\boldsymbol{q}_{upper}^*$ are the actual and desired upper body joint positions. $\sigma_{upper}$ is set to 0.35. Weight: 1.0.

    \item \textbf{Torso Yaw Tracking Reward} ($r_{\text{yaw}}$): This term encourages tracking of torso yaw orientation commands.
          \begin{equation*}
              r_{yaw} := \exp(-e_{yaw}^2/\sigma_{torso}^2),
          \end{equation*}
          where $e_{yaw}$ is the yaw orientation error. $\sigma_{torso}$ is set to 0.2. Weight: 0.25.

    \item \textbf{Torso Roll Tracking Reward} ($r_{\text{roll}}$): This term encourages tracking of torso roll orientation commands.
          \begin{equation*}
              r_{roll} := \exp(-e_{roll}^2/\sigma_{torso}^2),
          \end{equation*}
          where $e_{roll}$ is the roll orientation error. $\sigma_{torso}$ is set to 0.2. Weight: 0.25.

    \item \textbf{Torso Pitch Tracking Reward} ($r_{\text{pitch}}$): This term encourages tracking of torso pitch orientation commands with higher weight.
          \begin{equation*}
              r_{pitch} := \exp(-e_{pitch}^2/\sigma_{torso}^2),
          \end{equation*}
          where $e_{pitch}$ is the pitch orientation error. $\sigma_{torso}$ is set to 0.2. Weight: 0.5.

    \item \textbf{Center of Mass Tracking Reward} ($r_{\text{CoM}}$): This term maintains stability by keeping the center of mass near the support base.
          \begin{equation*}
              r_{CoM} := \exp(-\|\boldsymbol{p}_{CoM}^{xy} - \boldsymbol{p}_{feet}^{xy}\|_2^2/\sigma_{CoM}^2),
          \end{equation*}
          where $\boldsymbol{p}_{CoM}^{xy}$ is the horizontal center of mass projection and $\boldsymbol{p}_{feet}^{xy}$ is the midpoint between ankles. $\sigma_{CoM}$ is set to 0.2. Weight: 0.5.

    \item \textbf{Termination Reward}: This term penalizes episode termination.
          \begin{equation*}
              r_{ter} := -200.0 \cdot \mathbb{I}_{\text{terminated}}
          \end{equation*}
          where $\mathbb{I}_{\text{terminated}}$ is 1 if the episode terminates, otherwise 0.

    \item \textbf{Z-axis Linear Velocity Reward}: This term penalizes the robot for moving along the z-axis.
          \begin{equation*}
              r_{z} := -1.0 \cdot (v_z)^2
          \end{equation*}
          where $v_z$ is the z-axis linear velocity.

    \item \textbf{Energy Reward}: This term penalizes output torques to reduce energy consumption.
          \begin{equation*}
              r_{e} := -0.001 \cdot \sum_i |\tau_i \cdot \dot{q}_i|
          \end{equation*}
          where $\tau$ represents the joint torques and $\dot{q}$ represents the joint velocities.

    \item \textbf{Joint Acceleration Reward}: This term penalizes excessive joint accelerations to promote smooth motions.
          \begin{equation*}
              r_{ja} := -2.5 \times 10^{-7} \cdot \|\ddot{q}\|_2^2
          \end{equation*}
          where $\ddot{q}$ represents the joint accelerations of the configured joints.

    \item \textbf{Action Rate Reward}: This term penalizes rapid changes in actions to encourage smooth control.
          \begin{equation*}
              r_{ar} := -0.1 \cdot \|a_t - a_{t-1}\|_2^2
          \end{equation*}
          where $a_t$ represents the current action and $a_{t-1}$ represents the previous action.

    \item \textbf{Base Orientation Reward}: This term penalizes non-flat base orientation to maintain an upright posture.
          \begin{equation*}
              r_{ori} := -5.0 \cdot (\text{roll}^2 \cdot \text{mask}_{roll} + \text{pitch}^2 \cdot \text{mask}_{pitch})
          \end{equation*}
          where $\text{mask}_{roll}$ and $\text{mask}_{pitch}$ are adaptive masks based on torso command magnitudes.

    \item \textbf{Joint Position Limit Reward}: This term penalizes joint positions that exceed their soft limits.
          \begin{equation*}
              r_{\text{jpl}} := -2.0 \cdot \sum_i \max(|q_i| - q_{i,\text{limit}}, 0)
          \end{equation*}
          where $q_i$ represents the position of joint $i$, and $q_{i,\text{limit}}$ is the soft limit.

    \item \textbf{Joint Effort Limit Reward}: This term penalizes excessive torques on waist joints.
          \begin{equation*}
              r_{\text{jel}} := -2.0 \cdot \sum_i \max(|\tau_i| - 0.999 \cdot \tau_{i,\text{max}}, 0)
          \end{equation*}
          where $\tau_i$ is the torque and $\tau_{i,\text{max}}$ is the maximum torque limit.

    \item \textbf{Joint Deviation Reward}: This term penalizes joint positions that deviate from their default positions.
          \begin{equation*}
              \begin{aligned}
                  r_{jd} := & -0.15 \cdot \sum_i |q_i - q_{i,\text{default}}| \\
                            & -0.3 \cdot \sum_j |q_j - q_{j,\text{default}}|
              \end{aligned}
          \end{equation*}
          where $i$ represents hip yaw and ankle roll joints, and $j$ represents hip roll joints.

    \item \textbf{Feet Air Time Reward}: This term rewards appropriate stepping behavior for bipedal locomotion.
          \begin{equation*}
              r_{\text{fat}} := 0.3 \cdot \min(t_{\text{air}}, 0.4)
          \end{equation*}
          where $t_{\text{air}}$ is the air time when exactly one foot is in contact and velocity command is above 0.1 m/s.

    \item \textbf{Feet Slide Reward}: This term penalizes feet sliding during ground contact.
          \begin{equation*}
              r_{\text{sl}} := -0.25 \cdot \sum_i \|v_{i,xy}\|_2 \cdot \mathbb{I}(\text{contact}_i)
          \end{equation*}
          where $v_{i,xy}$ is the horizontal velocity of foot $i$, and $\mathbb{I}(\text{contact}_i)$ indicates if the foot is in contact.

    \item \textbf{Feet Force Reward}: This term encourages maintaining appropriate ground reaction forces.
          \begin{equation*}
              r_{\text{ff}} := -3 \times 10^{-3} \cdot \sum_i \min(\max(f_{z,i} - 500, 0), 400)
          \end{equation*}
          where $f_{z,i}$ is the vertical ground reaction force on foot $i$.

    \item \textbf{Feet Stumble Reward}: This term penalizes lateral forces that indicate stumbling.
          \begin{equation*}
              r_{\text{fs}} := -2.0 \cdot \sum_i \mathbb{I}(\|f_{xy,i}\|_2 > 5|f_{z,i}|)
          \end{equation*}
          where $f_{xy,i}$ represents the horizontal ground reaction forces.

    \item \textbf{Flying State Reward}: This term penalizes the robot when it is airborne.
          \begin{equation*}
              r_{\text{fly}} := -1.0 \cdot \mathbb{I}(\text{all feet off ground})
          \end{equation*}

    \item \textbf{Undesired Contacts Reward}: This term penalizes undesired contacts with the environment.
          \begin{equation*}
              r_{\text{uc}} := -1.0 \cdot \sum_{i \in \mathcal{C}} \mathbb{I}\left( \| \mathbf{F}_i \|_2 > 1.0 \right)
          \end{equation*}
          where $\mathcal{C}$ represents the set of contact points excluding ankle contacts.

    \item \textbf{Ankle Orientation Reward}: This term penalizes excessive ankle roll orientations.
          \begin{equation*}
              r_{\text{ankle}} := -0.5 \cdot \sum_i \|\text{gravity}_{xy,i}\|_2^2
          \end{equation*}
          where $\text{gravity}_{xy,i}$ is the projected gravity vector in each ankle frame.

\end{itemize}

\subsection{\ULC Hyperparameters}
We illustrate the hyperparameters of \ULC in Table~\ref{tab:hyperparameters}.

\begin{table}[h]
    \centering
    \small
    \begin{tabular}{l r}
        \toprule
        \textbf{Parameter}                 & \textbf{Value} \\
        \midrule
        Number of Environments             & 8192           \\
        Training Iteration                 & 10000          \\
        Environment Steps                  & 24             \\
        Number of Training Epochs          & 5              \\
        Mini Batch Size                    & 4              \\
        Max Clip Value Loss                & 0.2            \\
        Discount Factor                    & 0.99           \\
        GAE discount factor                & 0.95           \\
        Entropy Regularization Coefficient & 0.006          \\
        Learning rate                      & 1.0e-3         \\
        Schedule                           & adaptive       \\
        Desired KL                         & 0.01           \\
        Max Grad Norm                      & 1.0            \\
        Value Loss Coefficient             & 1.0            \\
        Observation History Length         & 6              \\
        Action Scale                       & 0.25           \\
        Episode Length                     & 20.0 s         \\
        Simulation Timestep                & 0.005 s        \\
        Control Decimation                 & 4              \\
        \bottomrule
    \end{tabular}
    \caption{Hyperparameters of \ULC.}
    \label{tab:hyperparameters}
\end{table}

\subsection{Architecture Details}
\label{subsec:append_arch}

Table~\ref{tab:network_arch} illustrates the network architecture of \ULC.

\subsection{\ULC Training Curves}
\begin{figure}[h]
    \vspace{5pt}
    \centering
    \includegraphics[width=1.0 \linewidth]{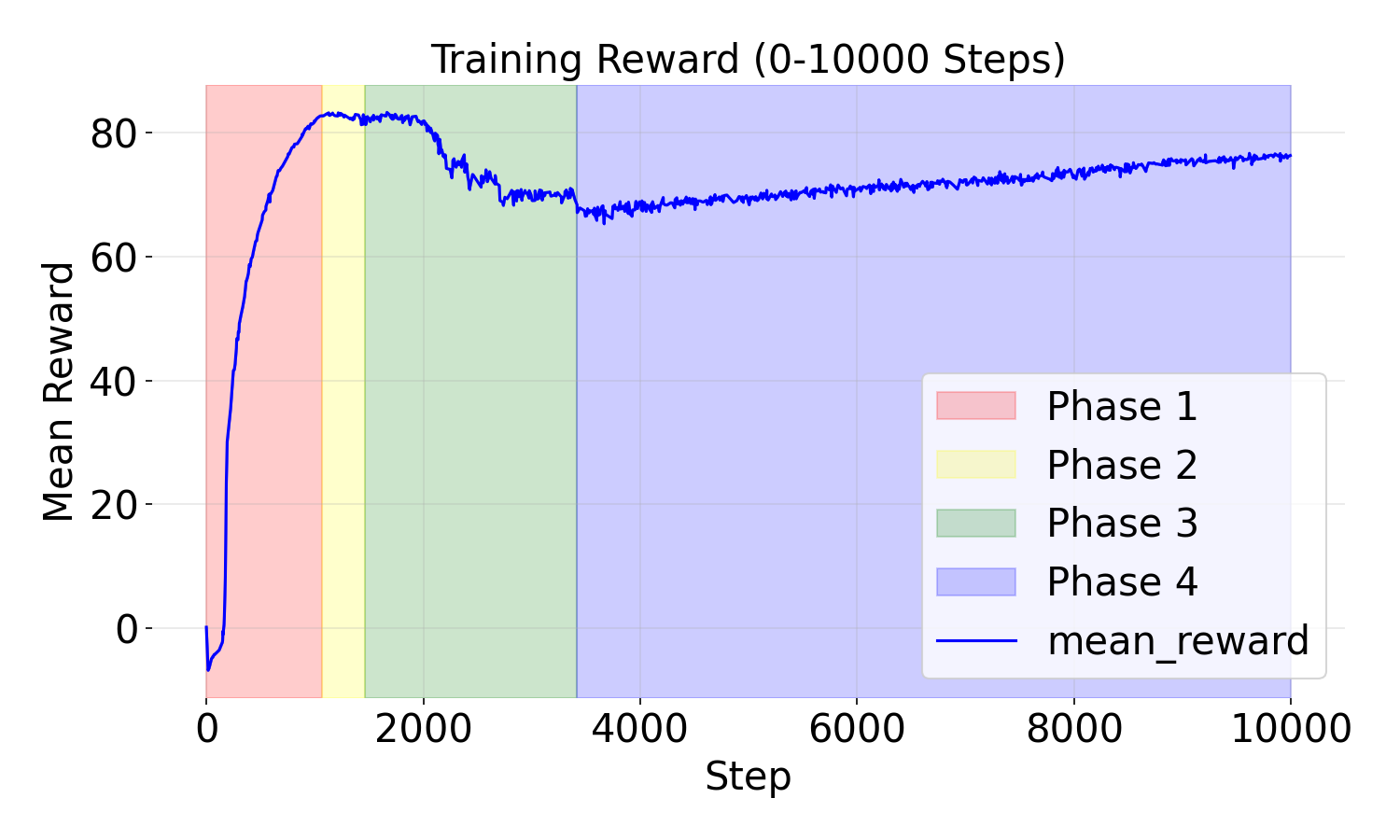}
    \caption{Reward convergence curve during \ULC training (first 10,000 iterations), showing four distinct stages of curriculum progression.}
    \label{fig:training_curves}
\end{figure}

We illustrate the training curves of \ULC in Figure~\ref{fig:training_curves},
showing the reward convergence over the first 10{,}000 training iterations. The
training process is divided into four distinct stages corresponding to the
sequential skill acquisition with adaptive curriculum:

\begin{enumerate}
    \item \textbf{Stage 1: Base velocity tracking initialization} ($T_1$ active, $\alpha_2 = 0$, $\alpha_3 = 0$): The policy learns fundamental locomotion skills by tracking base linear and angular velocity commands ($v_{xy}$, $\omega_z$) without additional curriculum constraints.

    \item \textbf{Stage 2: Height tracking curriculum activation} ($T_1$ and $T_2$ active, $\alpha_2$ increasing): Once the height curriculum advancement condition $\mathcal{C}_2$ is met, the base height tracking skill $T_2$ is activated. The curriculum parameter $\alpha_2$ progressively increases from 0 to 1.0, enabling the policy to learn pelvis height control ($h_{\text{pelvis}}$) while maintaining velocity tracking performance.

    \item \textbf{Stage 3: Upper body tracking curriculum activation} ($T_1$, $T_2$, and $T_3$ active, $\alpha_3$ increasing): When $\alpha_2 \geq 0.98$ and the upper body curriculum advancement condition $\mathcal{C}_3$ is satisfied, the torso and arm tracking skill $T_3$ is introduced. The curriculum parameter $\alpha_3$ increases from 0 to 1.0, allowing the policy to master torso orientation ($\boldsymbol{g}_{\text{torso}}$) and arm joint position ($\boldsymbol{g}_{\text{arms}}$) commands.

    \item \textbf{Stage 4: Full curriculum completion and convergence} ($T_1$, $T_2$, and $T_3$ active, $\alpha_2 = 1.0$, $\alpha_3 = 1.0$): All curriculum parameters reach their maximum values, enabling full command space exploration. The policy enters the final training phase with complete skill integration, leading to reward convergence across all tracking objectives.
\end{enumerate}

This staged progression ensures stable learning and prevents catastrophic
forgetting, with each stage building upon the competencies acquired in previous
phases.

\begin{table}[h]
    \centering
    \small
    \begin{tabular}{|l|c|}
        \toprule
        Component         & Configuration                    \\
        \midrule
        \multicolumn{2}{|c|}{Actor Network}                  \\
        \midrule
        Input Layer       & Observation (History × Features) \\
        Hidden Layer 1    & Linear(Input → 1024) + ELU       \\
        Hidden Layer 2    & Linear(1024 → 512) + ELU         \\
        Hidden Layer 3    & Linear(512 → 512) + ELU          \\
        Hidden Layer 4    & Linear(512 → 256) + ELU          \\
        Output Layer      & Linear(256 → 29)                 \\
        \midrule
        \multicolumn{2}{|c|}{Critic Network}                 \\
        \midrule
        Input Layer       & Observation (History × Features) \\
        Hidden Layer 1    & Linear(Input → 1024) + ELU       \\
        Hidden Layer 2    & Linear(1024 → 512) + ELU         \\
        Hidden Layer 3    & Linear(512 → 512) + ELU          \\
        Hidden Layer 4    & Linear(512 → 256) + ELU          \\
        Output Layer      & Linear(256 → 1)                  \\
        \midrule
        \multicolumn{2}{|c|}{Policy Distribution}            \\
        \midrule
        Distribution Type & Gaussian                         \\
        Initial Noise Std & 1.0                              \\
        Noise Type        & log                              \\
        \bottomrule
    \end{tabular}
    \caption{\ULC network architecture details.}
    \label{tab:network_arch}
\end{table}

\end{document}